\documentclass[10pt,twocolumn,letterpaper]{article}

\usepackage[pagenumbers]{cvpr} 

\usepackage{graphicx}
\usepackage{times}
\usepackage{epsfig}
\usepackage{amsmath}
\usepackage{amssymb}
\usepackage{booktabs}
\usepackage{multirow}
\usepackage{verbatim}
\usepackage{appendix}
\usepackage{subcaption}
\usepackage{enumitem}
\usepackage{comment}
\usepackage[accsupp]{axessibility} 
\usepackage[dvipsnames]{xcolor}
\usepackage[T1]{fontenc}
\usepackage[pagebackref,breaklinks,colorlinks]{hyperref}

\usepackage[capitalize]{cleveref}
\crefname{section}{Sec.}{Secs.}
\Crefname{section}{Section}{Sections}
\Crefname{table}{Table}{Tables}
\crefname{table}{Tab.}{Tabs.}

\def\cvprPaperID{6242} 
\def\confName{CVPR}
\def\confYear{2022}

\newcommand\blfootnote[1]{%
  \begingroup
  \renewcommand\thefootnote{}\footnote{#1}%
  \addtocounter{footnote}{-1}%
  \endgroup
}

\begin{document}

\title{STCrowd: A Multimodal Dataset for Pedestrian Perception in Crowded Scenes}

\author{\textbf{Peishan Cong}$^{1}$, \textbf{Xinge Zhu}$^{2}$, \textbf{Feng Qiao}$^{3}$, \textbf{Yiming Ren}$^{1}$, \textbf{Xidong Peng}$^{1}$, \textbf{Yuenan Hou}$^{4}$, \\
\textbf{Lan Xu}$^{1,7}$, \textbf{Ruigang Yang}$^{5}$, \textbf{Dinesh Manocha}$^{6}$, \textbf{Yuexin Ma}$^{1,7\dagger}$\\
$^{1}$ShanghaiTech University $^{2}$The Chinese University of Hong Kong\\
$^{3}$RWTH Aachen University $^{4}$Shanghai AI Laboratory\\
$^{5}$University of Kentucky $^{6}$University of Maryland at College Park\\
$^{7}$Shanghai Engineering Research Center of Intelligent Vision and Imaging\\
$^{1}${\tt\small \{congpsh, renym1, pengxd, mayuexin\}@shanghaitech.edu.cn }
}

\maketitle


\begin{abstract}

Accurately detecting and tracking pedestrians in 3D space is challenging due to large variations in rotations, poses and scales. The situation becomes even worse for dense crowds with severe occlusions. 
However, existing benchmarks either only provide 2D annotations, or have limited 3D annotations with low-density pedestrian distribution, making it difficult to build a reliable pedestrian perception system especially in crowded scenes.
To better evaluate pedestrian perception algorithms in crowded scenarios, we introduce a large-scale multimodal dataset, STCrowd. 
Specifically, in STCrowd, there are a total of 219 K pedestrian instances and 20 persons per frame on average, with various levels of occlusion.
We provide synchronized LiDAR point clouds and camera images as well as their corresponding 3D labels and joint IDs. STCrowd can be used for various tasks, including LiDAR-only, image-only, and sensor-fusion based pedestrian detection and tracking. We provide baselines for most of the tasks. In addition, considering the property of sparse global distribution and density-varying local distribution of pedestrians, we further propose a novel method, Density-aware Hierarchical heatmap Aggregation (DHA), to enhance pedestrian perception in crowded scenes.
Extensive experiments show that our new method achieves state-of-the-art performance for pedestrian detection on various datasets. \url{https://github.com/4DVLab/STCrowd.git}
\end{abstract}

\section{Introduction}
\label{sec:introduction}


\blfootnote{$\dagger$: Corresponding author.}

\begin{figure}[ht]
\centering
    \includegraphics[width=0.98\columnwidth,height=0.65\columnwidth]{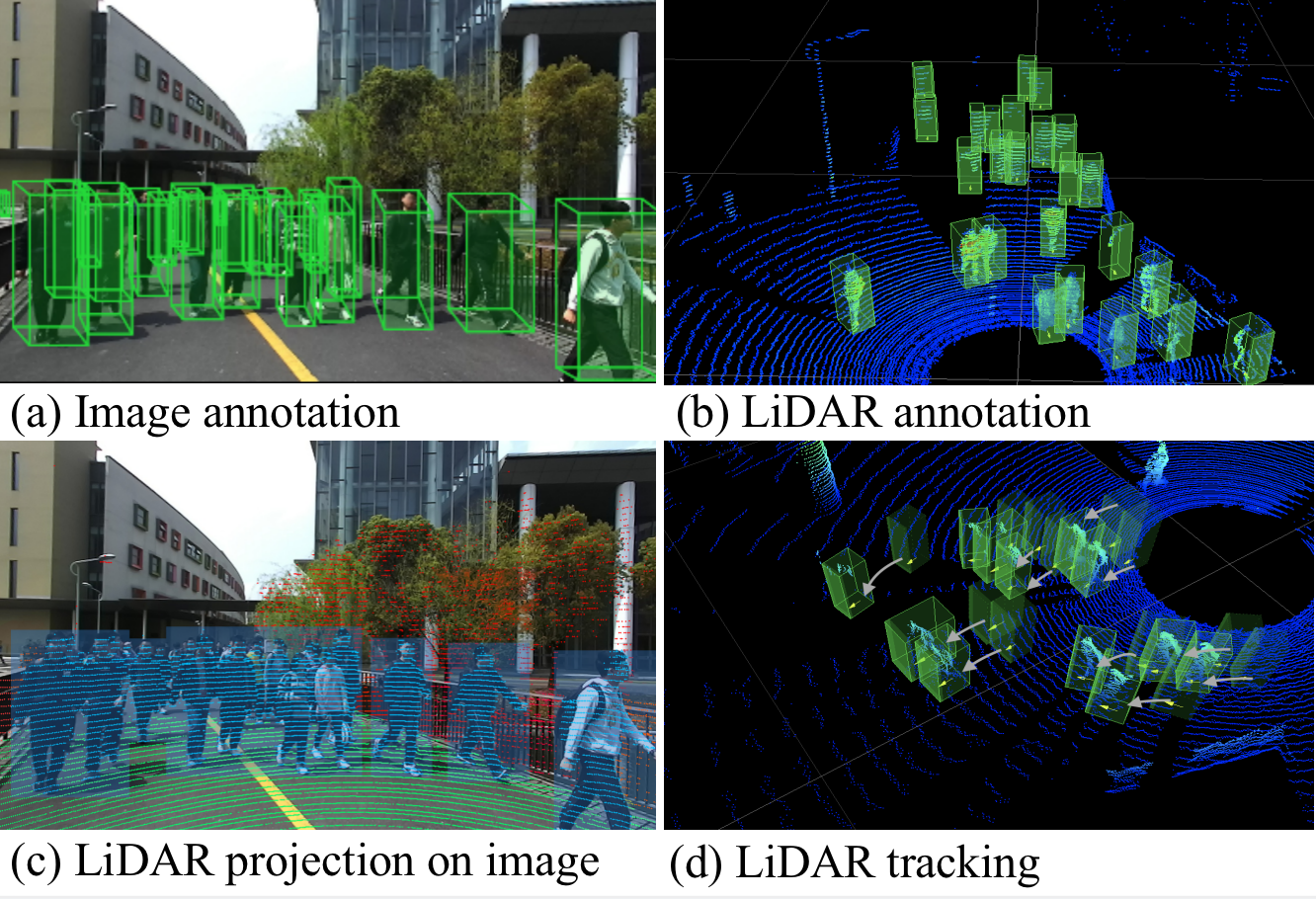}
\caption{STCrowd provides 2D/3D image annotations, 3D point cloud annotations, and joint annotations for consecutive frames. Note that STCrowd contains a large quantity of crowded scenes with severe occlusions, which pose great challenges to pedestrian detection and tracking.}
    \label{fig:4task}
\vspace{-3ex}
\end{figure}


Accurate pedestrian perception in 3D space plays a crucial role in thorough scene understanding. Many applications also benefit from reliable and accurate pedestrian perception~\cite{zhu2021cylindrical,Caesar2020nuScenesAM,bergmann2019tracking}, including surveillance, serving robots, autonomous driving, \etc.
%
%
However, pedestrian perception is intractable for three reasons. First, pedestrians are not rigid bodies and they can have various poses. Second, humans are relatively small for sensors to capture compared with other agents, such as vehicles. For instance, in LiDAR point clouds, pedestrians in the distance are usually represented as a few sparse points. Third, people tend to congregate when walking, which makes the detection of each individual person harder. Occlusion in crowded scenarios is a thorny problem for pedestrian perception. 

Many datasets~\cite{dollar2011pedestrian,hwang2015multispectral,zhang2017citypersons,zhang2019widerperson,Neumann2018NightOwlsAP,shao2018crowdhuman,Cordts2016TheCD,Che2019D2CityAL,Geiger2013VisionMR,Caesar2020nuScenesAM,Sun2020ScalabilityIP} have been collected to accelerate the research on the pedestrian perception field. Previous pedestrian perception datasets can be classified into two groups: image-based pedestrian datasets and multimodal traffic datasets. The former~\cite{dollar2011pedestrian,hwang2015multispectral,zhang2017citypersons,zhang2019widerperson,Neumann2018NightOwlsAP,shao2018crowdhuman,Cordts2016TheCD,Che2019D2CityAL,Yu2020BDD100KAD,chandra2019densepeds,chandra2021meteor,chandra2019traphic} focus on pedestrian detection and tracking on 2D images and merely provide 2D bounding box annotations, which is insufficient for deep models to infer accurate 3D positions of the pedestrians. The urgent demand for precise pedestrian perception in 3D space has given rise to a suit of 3D annotated datasets~\cite{Geiger2013VisionMR,Caesar2020nuScenesAM,Sun2020ScalabilityIP,Patil2019TheHD,Zhu2020A3DA3,Chang2019Argoverse3T,lyftl5}. 
However, these datasets all focus on the traffic scenes, where most objects on the roads are vehicles and pedestrians are distributed sparsely, which limits the exploration and evaluation of learning-based perception methods, especially for crowded scenes. 

Specifically targeting 3D pedestrian perception in challenging crowded scenarios, we introduce a large-scale multimodal dataset, STCrowd, with manually labeled 3D annotations for both images and point clouds. There are a total number of 219 K pedestrian instances in STCrowd with 20 persons per frame on average and more than 30 persons per frame in extremely crowded scenes. Specifically, there are 8 pedestrians in 5 meters on average centered on each person, which is much denser than contemporary 3D detection benchmarks, \eg, nuScenes~\cite{Caesar2020nuScenesAM} and KITTI~\cite{Geiger2013VisionMR}. Due to the lack of crowded 3D pedestrian datasets, perception algorithms always suffer from severe occlusions when dealing with crowded scenarios. STCrowd is very useful for exploring more effective methods and testing their robustness. In addition, we capture the data in 9 different scenes, covering different weather, light conditions and road conditions. With rich annotations, STCrowd is applicable for different tasks, including LiDAR-only, image-only, and sensor-fusion based detection, tracking and even trajectory prediction. We also provide baselines for most of the tasks in this paper to facilitate further research.

For LiDAR-captured outdoor scenes, pedestrians typically account for a small portion of the whole scene. For crowded scenarios, pedestrians gather together, which causes different degrees of occlusion and makes it difficult to distinguish each individual pedestrian accurately in the crowd. Considering the sparse global distribution and density-varying local distribution of the pedestrians, we propose a novel method, Density-aware Hierarchical heatmap Aggregation (DHA), to enhance pedestrian perception especially in crowded scenes. Specifically, DHA is comprised of the spatial attention module and the hierarchical heatmap aggregation module. The former makes the network focus on the pertinent foreground regions and the latter helps distinguish individuals in density-varying scenes via multi-level heatmaps. We evaluate our method on STCrowd and achieve state-of-the-art performance. The extension of the proposed DHA to tracking problem and ablation studies on various backbones also demonstrate its effectiveness and good generalization capability.
Our contribution is summarized as below:

\noindent 1) We propose a large-scale multimodal pedestrian-oriented dataset in crowded scenarios with 3D manual annotations. High-density distributions of pedestrians result in severe occlusion, which bring challenges for accurate perception. 

\noindent 2) Our dataset can be used for various tasks, including LiDAR-only, image-only, and sensor-fusion based pedestrian detection and tracking. We provide baselines and metrics for most of the tasks to facilitate further research.

\noindent 3) We propose a novel method to enhance LiDAR-based pedestrian perception in crowded scenes and achieve state-of-the-art performance on the STCrowd benchmark.

\section{Related Work}
\label{sec:relatedwork}

\noindent \textbf{Image-based Datasets}
Many datasets have been proposed over the last decade for pedestrian detection. Early datasets, like INRIA~\cite{dalal2005histograms}, ETH~\cite{ess2008mobile}, TUD-Brussels~\cite{wojek2009multi}, and Daimler~\cite{enzweiler2008monocular} are too small for the training and generalization of deep learning-based methods. Caltech~\cite{dollar2011pedestrian} is a widely-used pedestrian dataset with plenty of annotations. After that, more and more datasets were proposed for boosting data-driven human detection techniques, including KAIST~\cite{hwang2015multispectral}, CityPersons~\cite{zhang2017citypersons}, WiderPerson~\cite{zhang2019widerperson}, NightOwls~\cite{Neumann2018NightOwlsAP}, etc. Especially, CrowdHuman~\cite{shao2018crowdhuman} and DensePeds~\cite{chandra2019densepeds} provide many crowded scenes. In addition, there are also some datasets with traffic scenes containing pedestrians, such as CamVid~\cite{brostow2008segmentation}, Vistas~\cite{neuhold2017mapillary} Cityscapes~\cite{Cordts2016TheCD}, $D^2$-City~\cite{Che2019D2CityAL}, BDD100k~\cite{Yu2020BDD100KAD}, METEOR~\cite{chandra2021meteor}, etc. Note that all of them annotate pedestrians with 2D annotations, which is not applicable for real-world 3D perception.

\begin{figure*}[ht!]
\includegraphics[width=2.1\columnwidth,height=0.45\columnwidth]{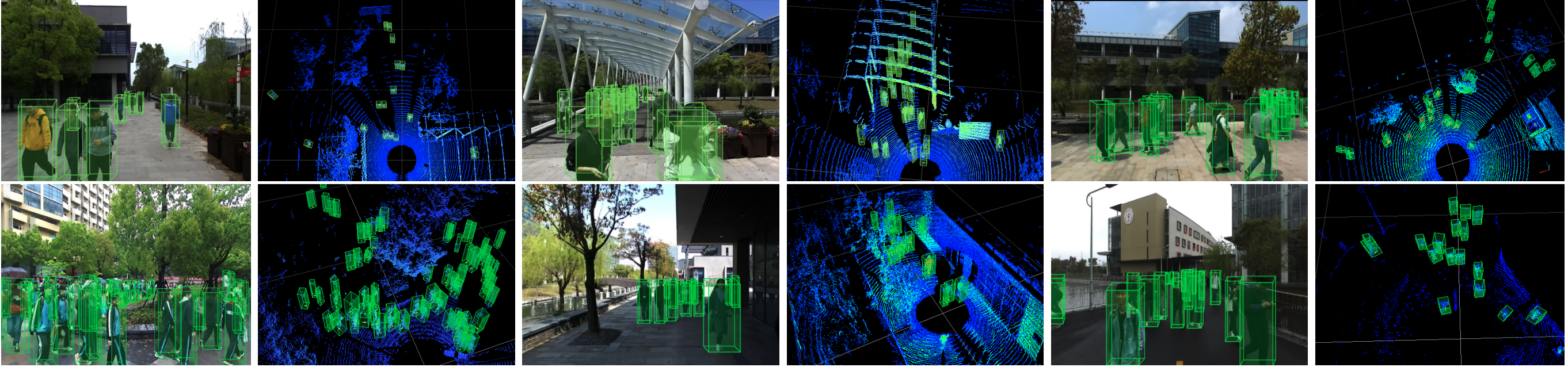}
	\caption{Scene examples of STCrowd with different backgrounds and weather, including clear weather (the first row), cloudy and rainy conditions (the second row). Note that the last figure shows the poor captured LiDAR point cloud due to the rainy conditions.}
	\label{fig:scenes}
\vspace{-2ex}
\end{figure*}

\noindent \textbf{Multimodal Datasets}
With the rapid development of autonomous driving, there is a growing demand for large-scale datasets with 3D annotations. Apolloscape Detection/Tracking~\cite{Ma2019TrafficPredictTP} is a challenging urban traffic scenes dataset. It only has 3D annotations for the LiDAR point cloud. Because almost all the autonomous vehicles have both cameras and LiDAR sensors for perception, there are many multimodal datasets containing synchronized images and LiDAR point cloud, such as KITTI~\cite{Geiger2013VisionMR}, nuScenes~\cite{Caesar2020nuScenesAM}, Waymo Open~\cite{Sun2020ScalabilityIP}, H3D~\cite{Patil2019TheHD}, A2D2~\cite{geyer2020a2d2}, KAIST~\cite{choi2018kaist}, A*3D~\cite{Zhu2020A3DA3}, Argoverse~\cite{Chang2019Argoverse3T}, Lyft L5~\cite{lyftl5}, etc. However, they all focus on the driving scenes, where vehicles account for most of the scene, and pedestrians are distributed sparsely. In fact, pedestrians have free rotations and diverse poses, and they are much smaller than vehicles, which dramatically increases the difficulty in the detection and tracking. Furthermore, high-density crowd scenes are much more challenging due to the existence of severe occlusion. Our dataset provides diverse scenes with various densities and distributions of pedestrians, which is significant for testing perception methods' generalization.

\noindent \textbf{3D Detection and Tracking} ~~~ LiDAR-only-based 3D detection methods aim to classify and locate the 3D bounding boxes in the given point cloud. Most of them ~\cite{zhu2021cylind,Yang2018PIXORR3,Qi2019DeepHV,Qi2020ImVoteNetB3, Vora2020PointPaintingSF, Zhou2019EndtoEndMF, Shi2021FromPT, Zhu2019ClassbalancedGA, Zhu2020SSNSS, Chen2020ObjectAH} first project the point cloud into a 3D or 2D representation, such as voxel and pillar. After that, the standard 2D convolution and 3D convolution are utilized to process these structured representations. Another group of existing methods~\cite{Shi2019PointRCNN3O, Yang20203DSSDP3, Yang2019STDS3} aims to process the point cloud in the raw data, which better preserves the 3D geometric information but has a high computational cost for the large-scale point clouds. Because of the  complementary  roles  of  point  clouds and  images, the LiDAR and camera fusion methods have gained much attention in recent years. PointPainting~\cite{Vora2020PointPaintingSF} makes the sensor fusion in the point level with a hard-association. PointAugmenting~\cite{Wang2021PointAugmentingCA} performs the point-wise fusion in the feature level. Furthermore, MV3D~\cite{Chen2017Multiview3O} and AVOD~\cite{Ku2018Joint3P} perform fusion at the region proposal level. Similarly, \cite{Liang2019MultiTaskMF, Liang2018DeepCF, Xie2020PIRCNNAE, Yoo20203DCVFGJ} project the point cloud onto {the} bird's eye view (BEV) and then fuse the image features in the BEV level. For the 3D tracking~\cite{hu2019joint}, existing methodologies for 2D tracking can be easily adapted to 3D space~\cite{bergmann2019tracking,wojke2017simple}. A common pipeline is to combine the 3D detectors and 3D Kalman filters to perform 3D tracking~\cite{chiu2020probabilistic,weng2019baseline}. However, most of them often struggle with the challenging crowded scenes with severe occlusions. Using spatial attention and hierarchical heatmaps, our method can focus on pedestrians in large-scale scenes and distinguish individuals well for density-varying crowds.
\section{Dataset}
STCrowd is collected by a 128-beam LiDAR and a monocular camera, which are synchronized and mounted at a fixed position on the vehicle. The detailed set-up of the sensors is shown in the supplementary materials. The annotated dataset is comprised of 84 sequences and the total number of frames is 10, 891. Each sequence contains a variable number of continuously recorded frames, ranging from 50 to 800. There are 219 K and 158 K instance-level bounding box annotations in point clouds and images, respectively. Joint annotations of point clouds and images in sequences are also provided. In particular, we get official permission for collecting the data and we protect personal privacy by blurring faces shown in images. The data annotation project involved 20 people with professional skills and took 960 man-hours effort. And we have two rounds of quality inspection for each batch of data.

\begin{table*}[h]
\centering
\caption{Comparison of STCrowd with popular multimodal datasets, where Fr denotes frames, PerFr denotes per frame of LiDAR point cloud, - represents unknown. Density-2/5/10 shows the average number of pedestrians within 2, 5, and 10 meters respectively centered on each pedestrian. Person/Range is the ratio of the number of pedestrians in each frame and the LiDAR scan diameter. The last three indicators measure the density of the dataset from different aspects. }
\label{compare}
\begin{tabular}{lllllllll}
              & LiDAR Fr  & 3D Boxes  & Beam   & Person Num & Person PerFr &  Person/Range & Density-2/5/10         \\\hline

PedX~\cite{kim2019pedx}          & 2.5k         & 0             & -        &  14k     &    5.6    & -               & -      \\\hline

Argoverse~\cite{Chang2019Argoverse3T} & 22k        & 993k             &32      &   110k     &    5       & -      & -             \\\hline
Lyft L5~\cite{lyftl5} & 46k         & 1.3M         &64   &   210k     &    4.6    & -     & -                 \\\hline

A*3D~\cite{Zhu2020A3DA3}           & 39k    & 230k             &64       & 20k        & 0.5  & -   & -                 \\\hline

KITTI~\cite{Geiger2013VisionMR}         & 15k      & 80k              &64       & 4.5k   & 0.3     &0.006   & 0.5/1.3/2.3        \\\hline
nuScenes~\cite{Caesar2020nuScenesAM}       & 40k        & 1.4M             &32      & 208k   & 5    &0.05 &   0.7/1.6/2.7              \\\hline
H3D~\cite{Patil2019TheHD}           & 27k         & 1M             &64        & 280k   & 10     &0.1        &1.5/4.0/7.2        \\\hline
Waymo Open~\cite{Sun2020ScalabilityIP}    & 230k        & 12M              &64         & 2.8M   & 12  & 0.16    & 1.0/2.9/5.6                    \\\hline
\hline
ours          & 11k        & 219k             &\textbf{128}    & 219k   & \textbf{20} & \textbf{0.4} & \textbf{2.4/8.0/15.8} \\ \hline
\end{tabular}
\vspace{-1ex}
\end{table*}

\begin{figure}
    \centering
    \includegraphics[width=0.98\linewidth,height=0.96\columnwidth]{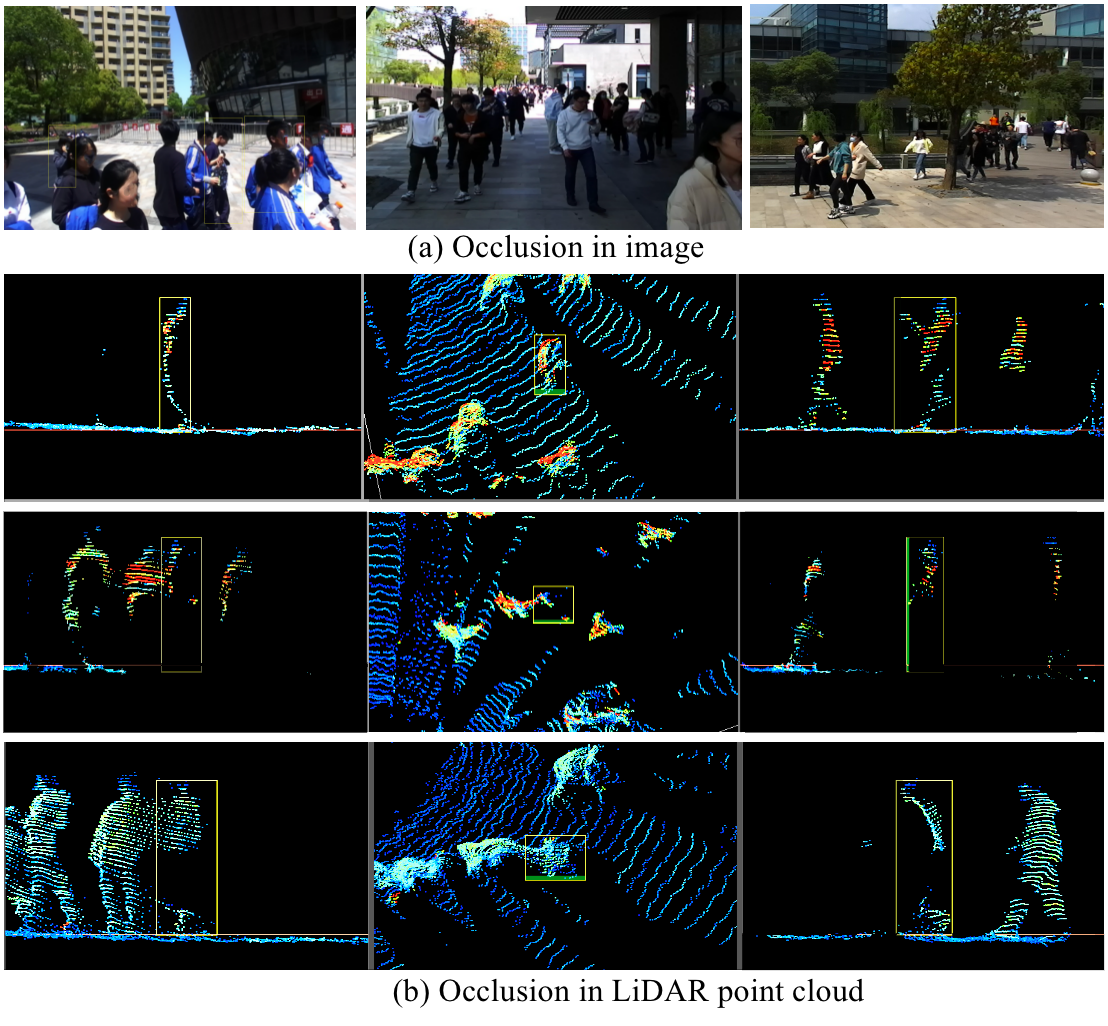}
    \caption{Occlusion cases in STCrowd. (a) shows occlusion cases in image. (b) demonstrates examples in LiDAR point cloud from different views (front, top, and side view) from left to right, in which severely occluded pedestrians are marked by yellow boxes.}
    \label{fig:occlusion}
    \vspace{-3ex}
\end{figure}

\subsection{Characteristics}

\noindent\textbf{Diverse scenes and weather.} We collect data in different scenes and weather conditions as shown in Figure.~\ref{fig:scenes}. Our scenes include rich background with bridges, trees, buildings and designed architectures. Unlike traffic scenes in which pedestrians gather around junctions or roadsides, the distribution of pedestrians in our scenes is more diverse. The weather also varies to include clear, cloudy and rainy days. Different lighting conditions will influence the color information of the image, and the rainy or wet conditions will affect the reflection from the LiDAR sensor, resulting in fewer points on objects and the background (eg. for the last picture shown in Figure.~\ref{fig:scenes}, it is clear to see that the captured points are very limited), which is challenging for perception algorithms. 



\noindent\textbf{Diverse crowd densities.} STCrowd contains crowd scenarios of various densities, which are divided into four levels, including fewer than 10, 10 $\sim$ 20, 20 $\sim$ 30, and more than 30 pedestrians. High-density is the most important characteristic of STCrowd. Table.~\ref{compare} shows the comparison with related datasets, which are widely-used for the 3D perception of large-scale outdoor scenes. STCrowd is notable on three evaluation values. The first is the number of pedestrians per frame. Our dataset has 20 pedestrians on average, which obviously exceeds others. The second is the ratio of pedestrians and the scene range captured by LiDAR, which can show the density of the distribution of pedestrians in the whole scene. Statistics show that the value of our dataset is 2.5 times the density in Waymo and more than 4 times others. The last evaluation is to illustrate the degree of the crowd gathering by computing the average number of pedestrians in 2, 5, and 10 meters centered on each pedestrian. There are 2.4, 8, and 15.8 persons under such measurements for our dataset, which reveals the local high-density characteristic of STCrowd. When people gather, point clouds of different instances always stick with each other, which makes it difficult for perception methods to distinguish individuals accurately.

\begin{figure}
    \centering
         \includegraphics[width=0.95\columnwidth,height=0.43\columnwidth]{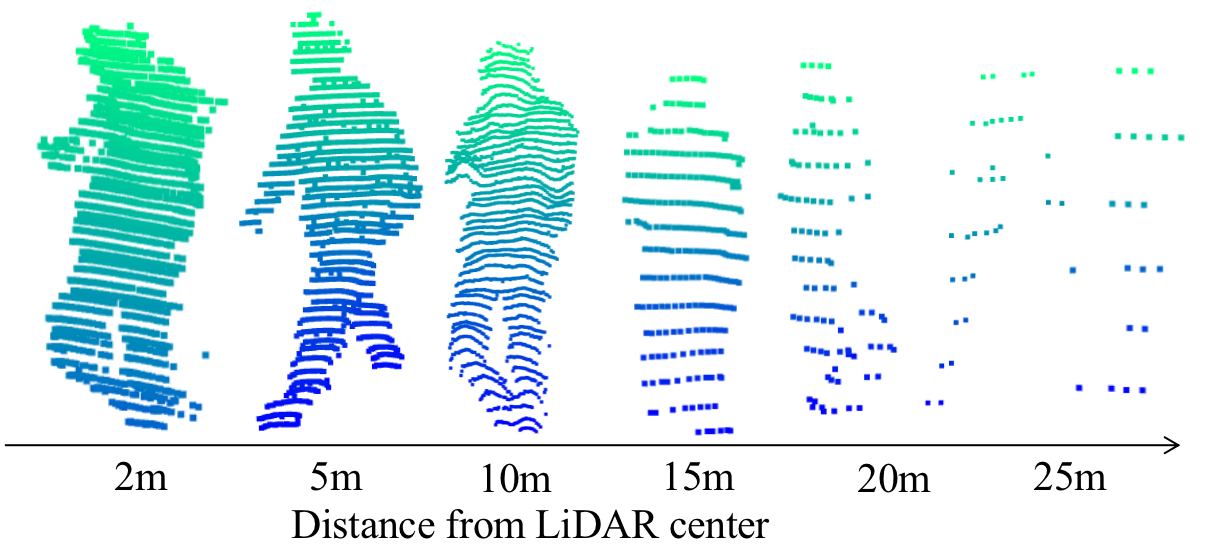}
    \caption{Wide-span point cloud densities for instances with the distance to LiDAR sensor increasing. }
    \label{fig:Wide-span density}
      \vspace{-3ex}  
\end{figure}


Dense crowds lead to severe occlusions in both images and LiDAR point cloud. As shown in Figure.~\ref{fig:occlusion}, many pedestrians only have a partial body or only one arm or a head, which makes accurate perception difficult due to limited partial features. We annotate occlusion labels (from 0 to 2) for each pedestrian, measuring how much it is occluded, where 0, 1, and 2 denote none of the body, no more than half the body and over half the body occluded, respectively. Hierarchically dividing the dataset according to occlusion situations can help test the performance of methods in dealing with challenging cases.

Besides various scene-level densities, we also demonstrate diverse instance-level densities. As shown in Figure.~\ref{fig:Wide-span density}, the point clouds of pedestrians become sparser as the distance to the LiDAR sensor increasing. Long-distance instances are hard to detect because the shape and scale information may loss on sparse points.

\begin{figure}
    \centering
    \includegraphics[width=0.99\columnwidth,height=1\columnwidth]{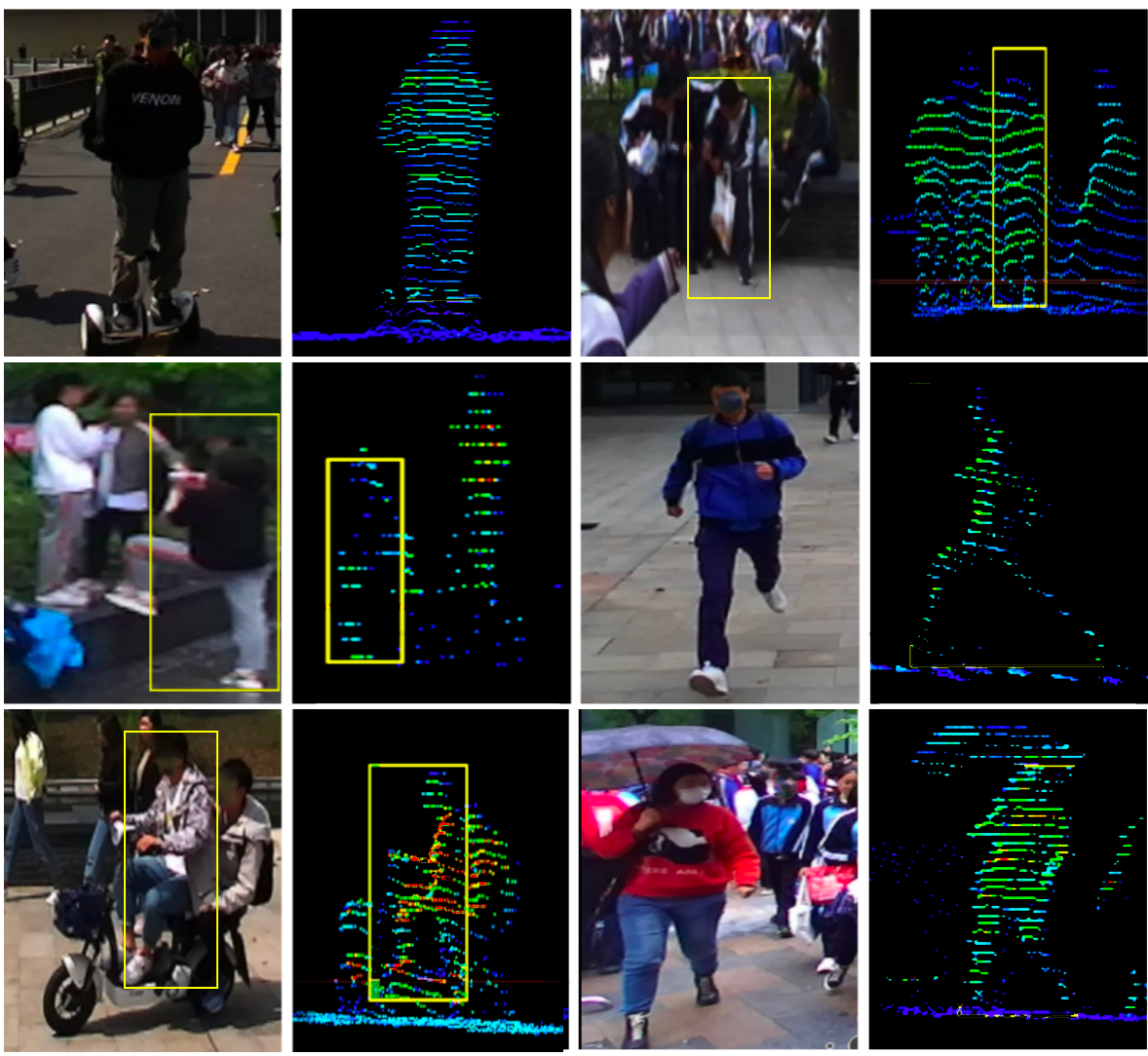}
    \caption{Diverse human poses in STCrowd. The same pedestrian in the image and the point cloud is marked by yellow boxes.}
    \label{fig:pose}
        \vspace{-2ex}
\end{figure}

\noindent\textbf{Diverse human poses.} Our dataset has a diversity of human poses. Figure \ref{fig:pose} shows some examples, like walking in person or in group, running, taking bicycles, taking balance cars, sitting, holding an umbrella, \etc. The diversity in pedestrian poses further increases the difficulty of accurate perception.
	
\subsection{Annotations}
We provide high-quality manually labeled ground truth for both LiDAR point clouds and images. For annotations in point clouds, we labeled each pedestrian using a 3D bounding box $ (x,y,z,l,w,h,\theta)$, where $x,y,z$ denotes the center coordinates and $l,w,h$ are the length, width, and height along the x-axis,y-axis and z-axis, respectively. Pedestrians with fewer than 15 points in the LiDAR point cloud are not annotated. For annotations of images, besides 3D bounding box, we also label the 2D bounding box with $x,y,w,h$ for general 2D detection and tracking. For the objects captured by both the camera and LiDAR, we annotate the joint ID in sequences, which facilitates tracking and sensor-fusion tasks. The frequency of our annotation is 2.5HZ. We also provide annotations for the level of density and occlusion, which is mentioned above.



\subsection{Tasks \& Metrics}
Our multi-modal dataset supports detection, tracking, and prediction tasks. We give evaluation metrics in this section to provide benchmarks on our dataset.
\subsubsection{Detection Metric}\label{apm}
\noindent\textbf{Average Precision metric.}
Following~\cite{Caesar2020nuScenesAM}, we use Average Precision (AP) metric with the 3D center distance threshold. 
For crowded scenes, the distance thresholds are chosen from $D = \{0.25,0.5,1\}$ meters and the mean Average Precision (mAP) is calculated by:
 $$mAP = \frac{1}{|D|} \sum_{d \in D} AP_d$$

\noindent\textbf{Average Recall with different occlusion levels.}
In addition to AP, for crowded scenes, the performance on occluded instances are also considered, and we calculate the average recall with different center distance thresholds $D = \{0.25,0.5,1\} $ for different levels of occlusion $i$: 
 $$AR_i = \frac{1}{|D|} \sum_{d \in D} Recall_{i,d}, i \in \{0,1,2\}$$

\subsubsection{Tracking Metric}
\noindent\textbf{MOTA}
We use traditional Multi-Object Tracking Accuracy (MOTA) to measure the tracking result:
$$MOTA = 1-(FP+IDS+FN)/GT$$ where FP and FN are false positive and false negative, IDS denotes the false ID matching for tracking in different time-steps, and GT is the number of ground truth tracked instances.

\noindent\textbf{ML \& MT}
ML(mostly loss) is the proportion of successful track matching of ground truth in less than 20\% of the time in all tracking targets. MT(mostly track) is the proportion of successful track matching of ground truth in more than 80\% of the time in all tracking targets.

\subsubsection{Prediction Metric}
\noindent\textbf{FDE \& MDE} Final displacement error (FDE) is the Euclidean distance between the predicted output and the ground truth at the last time step and Mean displacement error (MDE) is the average Euclidean distance between the predicted output and the ground truth for each time step.
\section{DHA: Density-aware Hierarchical Heatmap Aggregation}
\label{sec:methodology}

For pedestrian detection in crowded scenarios, pedestrians are always walking together, resulting in local high density and occlusion in both the point cloud and the image. Moreover, the density distribution and pose types of pedestrians vary. These challenges narrow down the capability of existing methods for accurate pedestrian detection in such conditions. To tackle these problems, we propose the density-aware hierarchical heatmap aggregation (DHA) module shown in Figure.~\ref{fig:model}, which makes the model learn the attention on the location of individuals and produces multi-scale predictions covering regions with different densities, in which the proposed module could mitigate too much background influence and tackle the problem of pedestrian clustering at various densities levels.

In what follows, we give detailed explanations for these two main components, \ie, the spatial attention module focusing on the pertinent regions of pedestrians and the multi-level heatmap loss covering varying density conditions.
\subsection{Spatial Attention Module}

\begin{figure}
    \centering
    \includegraphics[width=0.92\columnwidth,height=0.65\columnwidth]{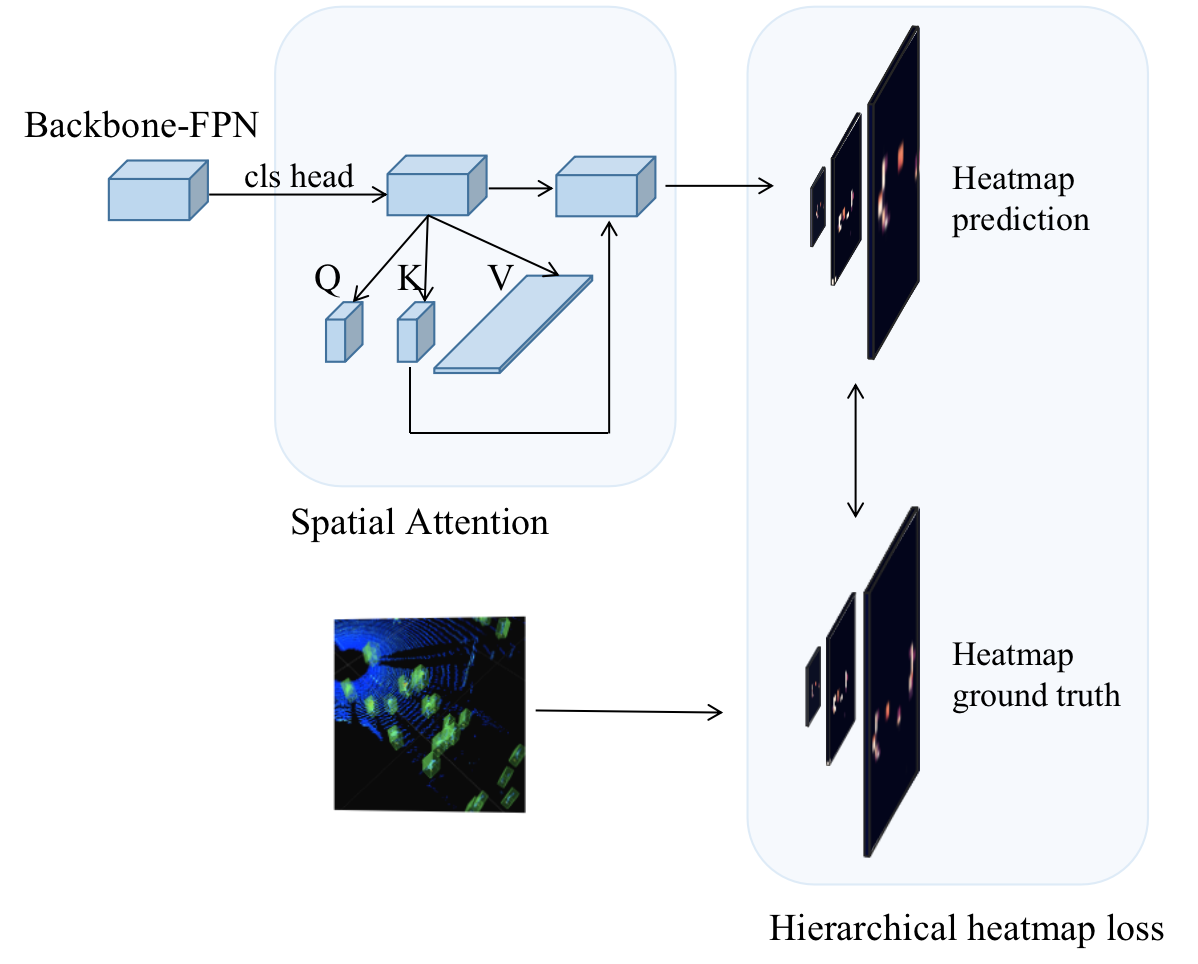}
    \caption{Density-aware hierarchical heatmap aggregation. We design the spatial attention with multi-level Gaussian score map supervision to tackle the problem of various density distribution and background influences, where it can also act as a plug-in for different backbones.}
    \label{fig:model}
    \vspace{-2ex}
\end{figure}

Crowds of pedestrians tend to be clustered, presenting locally high density and globally sparse distribution. Hence, attention to these foreground regions is crucial to the performance of pedestrian detection. To this end, we follow ~\cite{vaswani2017attention} to model the global attention with a transformer. As shown in Figure.~\ref{fig:model}, we apply the triplet $<$Query, Key, Value$>$ attention layer to extract the correlation among different locations and reweight these locations. For the final output $\bar{X}$, $$ \bar{X} = \text{softmax} (QK^T) \times V ,$$ where $Q$, $K$, $V$ denote the output of Query, Key and Value layer.


\begin{figure}
    \centering
    \includegraphics[width=0.95\columnwidth,height=0.63\columnwidth]{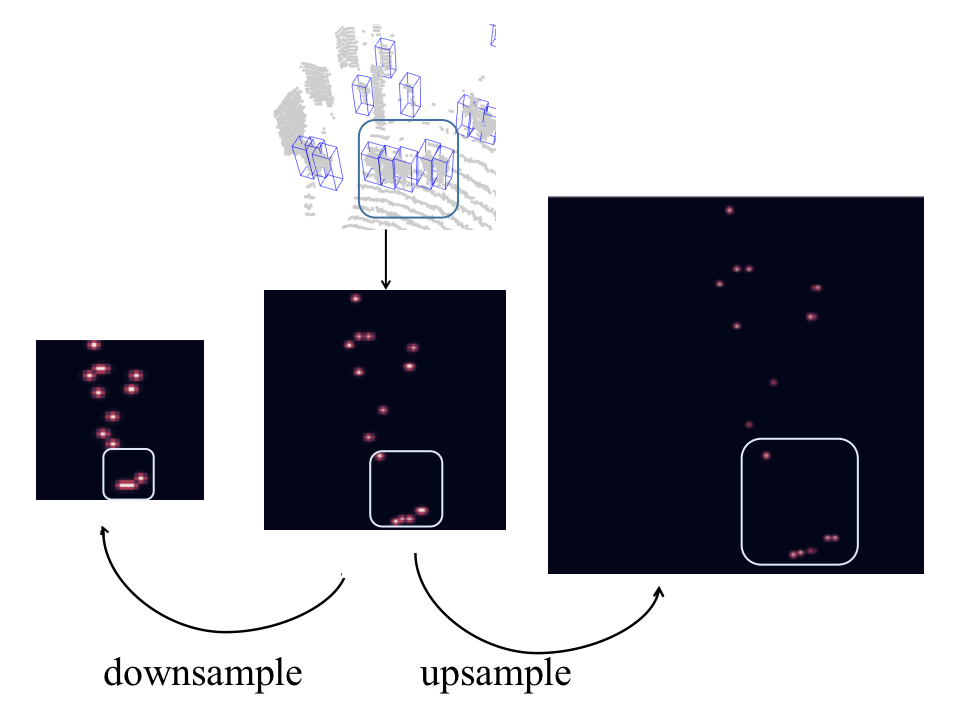}
    \caption{Hierarchical heatmaps. We design the spatial Hierarchical Gaussian score heatmap supervision to tackle the problem of various density distributions and background influences.
    The coarse-level heatmap is downsampled from the regular one, in which the positive regions occupy a larger portion targeting a balanced foreground / background ratio. For the fine-grained level heatmap, it can clearly distinguish the closer pedestrians for accurate one-to-one assignment (as shown in the rectangle).}
    \label{fig:mhl}
\vspace{-2ex}
\end{figure}

\subsection{Hierarchical Heatmaps}

Balanced positive and negative samples also have a great impact on high-performance 3D detection. However, unlike big objects (like truck covering a large portion of heatmap), pedestrians only occupy a limited space, resulting in most of the heatmap being zero (or negative region). The density-varying property also worsens the condition.
We thus introduce hierarchical heatmaps to construct the multi-level detection targets, where the coarse-level heatmap could balance the ratio of positive and negative samples and the fine-grained heatmap can better tackle the clustering pedestrians with an accurate one-to-one assignment (avoid one grid in the heatmap representing more than two persons).

Specifically, we use a modified CenterHead~\cite{Yin2020Centerbased3O} to classify and localize the pedestrians. The ground truth heatmap is a Gaussian map produced based on 3D centers of annotated bounding boxes. 
As shown in Figure~\ref{fig:mhl}, the features generated from Spatial attention module are first up-sampled to get the fine-grained feature maps and corresponding heatmaps, and then down-sampled to obtain the coarse-level features, where fine-grained level heatmap delivers an accurate one-to-one assignment for close and crowded pedestrians and coarse-level heatmap balances the ratio of foreground and background samples.
Gaussian focal loss is calculated on each pair of prediction and target heatmaps. 

With the cooperation of these two components, \ie, spatial attention module and hierarchical heatmaps, DHA handles the crowd scenarios with varying density well.

\section{Experiments}
\label{sec:experiments}

\begin{table*}[ht]
\centering
\caption{Benchmarks for image-only, LiDAR-only, and LiDAR-image-fusion-based 3D detection on validation set of STCrowd. AP($d$) denotes that different meters are used as matching thresholds of 3D center distance $d$. $AR_i$ represents the average recall on easy, moderate, and hard cases respectively with different occlusion levels $i$. }
\begin{tabular}{c|c|cccc|ccc}
\toprule
                 Methods   & Modality & AP(0.25) & AP(0.5) & AP(1.0) & mAP &  $AR_0$         & $AR_1$       & $AR_2$   \\\toprule
CenterNet~\cite{Zhou2019ObjectsAP}+ResNet18 &  RGB & 0.009    & 0.091   & 0.397   & 0.166&0.456 & 0.350 & 0.285 \\\hline
CenterNet~\cite{Zhou2019ObjectsAP}+ResNet101&  RGB & 0.056    & 0.112   & 0.486   & 0.203&0.478 & 0.361&  0.273 \\\hline
CenterNet~\cite{Zhou2019ObjectsAP}+DLA34 &  RGB & 0.041    & 0.200   & 0.467   & 0.236&0.578 & 0.451&  0.349 \\\midrule
\midrule
PointAugmenting~\cite{Wang2021PointAugmentingCA}  & RGB+LiDAR &0.483 &0.629 &0.649 &0.587 &0.932 &0.866 &0.800 \\\hline
PointPainting~\cite{Vora2020PointPaintingSF} & RGB+LiDAR &0.509 &0.638 &0.656 &0.601 &0.929 &0.867 &0.783 \\\midrule
\midrule
PointPillar~\cite{Lang2019PointPillarsFE} & LiDAR & 0.091    & 0.276   & 0.368  & 0.245 &0.576 & 0.399&  0.238  \\\hline
Pillar-Center \cite{Yin2020Centerbased3O}    & LiDAR  & 0.456    & 0.574   & 0.592   & 0.541  & 0.866 &0.811 & 0.706 \\\hline
Voxel-Center\cite{Yin2020Centerbased3O} & LiDAR & \textbf{0.505}    & 0.613   & 0.628   & 0.582 & 0.859       & 0.834 & 0.740\\\midrule
Ours& LiDAR & 0.498    & \textbf{0.667}   & \textbf{0.685}   & \textbf{0.617} & \textbf{0.902} & \textbf{0.873} & \textbf{0.782} \\\bottomrule
\end{tabular}
\label{tab:detection}
\end{table*}

\begin{table}[ht]
\caption{Mean Average Precision (mAP) for 3D pedestrian detection on Waymo ($W_{range}^{level}$)~\cite{Sun2020ScalabilityIP}, H3D~\cite{Patil2019TheHD}, and nuScenes~\cite{Caesar2020nuScenesAM} datasets.}
\vspace{-2ex}
\footnotesize
\centering
\begin{tabular}{c|c|c|c|c|c|c}
\toprule
             Dataset & $W_{30}^1$ & $W_{30}^2$ & $W_{50}^1$ & $W_{50}^2$ & H3D & nuScenes \\\toprule
Pillar-Center  & 0.697 &  0.652  &0.535   & 0.472& 0.478 &0.719\\\hline
Voxel-Center  & 0.701 & 0.649 & 0.598& 0.532& 0.595& 0.783 \\\midrule
Ours  &\textbf{0.726} &  \textbf{0.673} & \textbf{0.638}& \textbf{0.569}& \textbf{0.609} & \textbf{0.795}\\\bottomrule
\end{tabular}
\label{more result}
\end{table}

In this section, we first provide the detailed experimental setup, then evaluate extensive methods and our proposed DHA for 3D detection on STCrowd. Furthermore, we demonstrate the performance of DHA on 3D tracking and prove its generalization capability by ablation study. Finally, the benchmark of trajectory prediction on our dataset is provided to facilitate the research of crowd prediction. Moreover, we provide more analyses and qualitative results in the supplementary material.

\subsection{Baseline}
We present several popular baselines with different modalities for 3D detection. For image-based method, CenterNet~\cite{Zhou2019ObjectsAP} regresses 3D bounding boxes in the world-coordinate system from only monocular images. For LiDAR-only 3D detectors, anchor based and anchor-free methods are used to evaluate the performance on our dataset, including PointPillar\cite{Lang2019PointPillarsFE} and CenterPoint\cite{Yin2020Centerbased3O}. Furthermore, we also evaluate two point encoding methods for CenterPoint, \ie, Voxel-CenterPoint and Pillar-CenterPoint. Our method takes the Voxel-CenterPoint as the backbone and employs DHA as the classification head. For LiDAR-image-fusion-based 3D detectors, we show current SOTA PointPainting~\cite{Vora2020PointPaintingSF} and PointAugmenting~\cite{Wang2021PointAugmentingCA}, where pixel-wise prediction and pixel-wise image features are used to represent the image. Note that all the backbone for these fusion-based methods is the Voxel-CenterPoint.

\subsection{Implementation Details}

For experiments on STCrowd, we set the detection range to [0, 30.72m] for the X axis, [-20.48m, 20.48m] for Y axis, and [-4m, 1m] for Z axis. Voxel-CenterPoint employs a (0.12m, 0.16m, 0.2m) voxel size, and Pillar-CenterPoint and PointPillars utilize a (0.12m, 0.16m) grid. For the anchor-based method, the anchor is set as [0.57m, 0.6m, 1.7m] which is calculated as the average size of pedestrian 3D ground truth bounding boxes. For post-processing of detection results, we use a circle NMS method which keeps only one instance prediction within radiance fewer than 0.3m to reduce redundant bounding boxes and drops the predicted box which has fewer than 5 points.

\subsection{Results}
\noindent\textbf{LiDAR-only 3D detection}
We compare the results of our proposed method with existing anchor-based and anchor-free methods in Table.~\ref{tab:detection}. Specifically, the anchor-free methods perform much better than anchor-based methods. The various human poses and densities become a hindrance for anchor-based methods, making it difficult for these anchors to cover pedestrian locations well. Moreover, the proposed DHA achieves state-of-art performance compared with anchor-free methods, demonstrating that DHA can tackle the issue of density-varying and unbalanced samples well, while vanilla anchor-free methods do not well attend.

For crowded scenes, we show the result of average recall $AR_i$ (as shown in Section~\ref{apm}) for different occlusion levels in Table.\ref{tab:detection} to evaluate the performance when facing key challenges in pedestrian detection. Consistently, the proposed method achieves about $4\%$ and $6\%$ improvement for each level compared with the Voxel-CenterPoint and Pillar-CenterPoint backbone, respectively.
 
\begin{figure}
    \centering
    \includegraphics[width=0.99\linewidth,height=0.7\columnwidth]{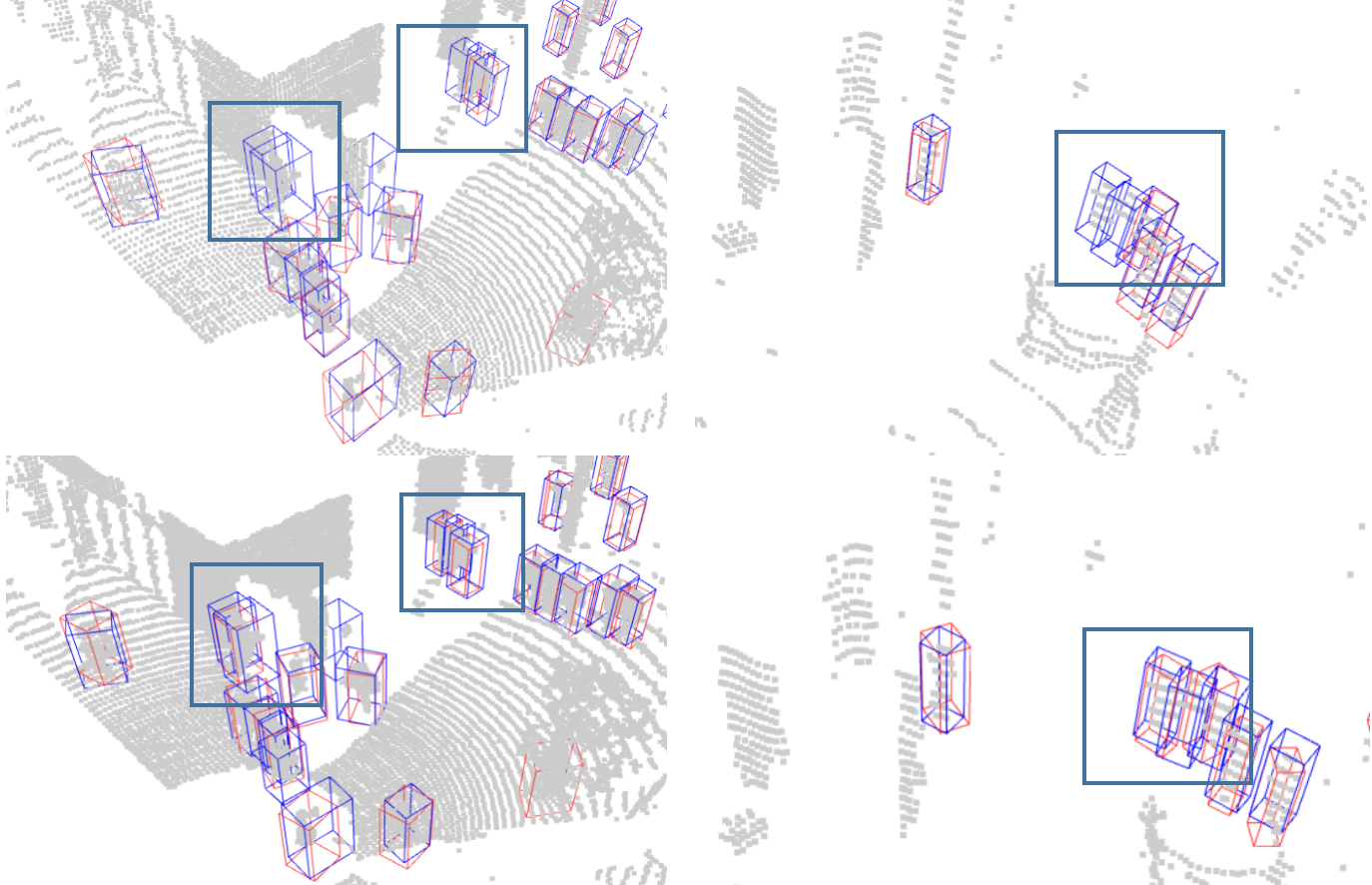}
    \caption{The detection visualization on crowded scenes. The first row is the prediction results from the baseline method~\cite{Yin2020Centerbased3O} and the bottom shows our results. The blue boxes are ground truth, and the red boxes are predictions. It can be found that for some crowded regions and pedestrians (as shown in rectangles), the baseline method often omits and mismatches, while our method achieves better results on such cases because the proposed DHA can focus more on the foreground and distinguish the crowded regions with fine-grained heatmaps.}
    \label{fig:exp-vis}
\end{figure}

 We further provide the visualization results for the detection task in crowded scenes in Figure.~\ref{fig:exp-vis}. It can be observed that in crowded scenes, our method performs much better, even covering these challenging pedestrians when they are extremely close to each other and occluded severely (as shown in rectangles), while the baseline cannot distinguish these boxes clearly and misses some dense predictions. 
 
 We also conduct experiments in Table.~\ref{more result} for 3D pedestrian detection on three other large-scale datasets. Obviously, our method consistently outperforms baselines (current SOTAs) on contemporary benchmarks, demonstrating good generalization capability of our method on 3D pedestrian detection. 

 \noindent\textbf{Multimodal fusion-based 3D detection}
As shown in Table.~\ref{tab:detection}, both fusion methods perform better than LiDAR-only baselines, which demonstrates the image features play a complementary role compared with LiDAR point clouds. Although our method only unitize LiDAR features, it is still comparable to sensor-fusion-based methods.

\noindent\textbf{Image-only 3D detection}
It is obvious that there is a big gap between the image-only method and others. Because it is difficult to estimate the depth information from monocular images and severe occlusions of the crowd in images make things worse. In this manner, LiDAR point clouds provide an appealing option and act as a complementary function to tackle these occlusions to some extend.

\subsection{Ablation studies}


\begin{table}[h]\small
	\centering
	\setlength{\tabcolsep}{0.6mm}
	\caption{Ablation studies for DHA block on different backbones on STCrowd validation set. 
	} \label{tab:Ablation1}
	\begin{tabular}{c|c|c|c|c}
		\toprule
		Pillar-Center & Voxel-Center & PointAugmenting & DHA & mAP \\
		\hline 
\textcolor{OliveGreen}{\checkmark}& && & 0.541\\		\hline 
\textcolor{OliveGreen}{\checkmark} && & \textcolor{OliveGreen}{\checkmark}  &	0.591\\ 	\hline 
&\textcolor{OliveGreen}{\checkmark}& && 0.582\\		\hline 

&\textcolor{OliveGreen}{\checkmark} &&\textcolor{OliveGreen}{\checkmark} &	0.617\\ 	\hline 
& &\textcolor{OliveGreen}{\checkmark} &  &  	0.587\\ 	\hline 
&&\textcolor{OliveGreen}{\checkmark} &\textcolor{OliveGreen}{\checkmark} &0.594\\	

\bottomrule 
	\end{tabular}

\end{table}

\begin{table}[ht]
\caption{Ablation study of DHA on STCrowd.}
\vspace{-2ex}
\centering
\begin{tabular}{c|c|c|c|c}
\toprule
             Methods & ours & w/o SAM  & w/o HH &  Voxel-Center     \\\toprule
mAP     &0.617  & 0.601   &  0.603 & 0.582  \\\hline

\end{tabular}
\label{ablation result}
\end{table}

First, we perform ablation experiments to investigate the generalization ability of the proposed modules on various backbones. The results on the validation set are reported in Table \ref{tab:Ablation1} and we test our DHA module on pillar-based and voxel-based LiDAR-only-based detection backbones. The results demonstrate that it consistently improves the performance with a large margin, \ie, 5\% and 3.5\%, respectively, compared to the baseline. We also test DHA on LiDAR-image-fusion-based detection backbone, PointAugmenting, and still get improvement (because image features already provide a remedy for the crowded scene, DHA only achieves a slight gain). We also conduct the ablation study for Spatial Attention Module (SAM) and Hierarchical Heatmaps (HH) (Table.~\ref{ablation result}). Results show the effectiveness of two modules in improving the performance.

\subsection{3D Tracking}
 For tracking tasks, we learn to predict a two-dimensional velocity estimation for each detected object as an additional regression output following the methodology of CenterPoint\cite{Yin2020Centerbased3O}. We mainly conduct experiments on CenterPoint~\cite{Yin2020Centerbased3O} with pillar and voxel representation, respectively. The proposed DHA module is also incorporated to investigate its generalization ability on the tracking task (Table.\ref{tracking result}). It can be found that our proposed DHA consistently achieves the better results.

\begin{table}[ht]
\caption{Results of LiDAR point cloud tracking on validation set of STCrowd. }
\centering
\begin{tabular}{lllll}
\toprule
             Methods     & MOTA~$\uparrow$  & MT~$\uparrow$ & ML~$\downarrow$ \\\toprule
Pillar-Center       & 0.245     &  0.295  & 0.102   \\\hline
Voxel-Center & 0.342 &   0.355      & \textbf{0.084}   \\\midrule
Ours & \textbf{0.368} &  \textbf{0.363}    &  0.086     \\\bottomrule
\end{tabular}
\label{tracking result}
\vspace{-2ex}
\end{table}

\subsection{Trajectory Prediction} As shown in Table.~\ref{predicition result}, we also provide the baselines of trajectory prediction on our crowd dataset, including popular vanilla-LSTM~\cite{greff2016lstm}, social-LSTM~\cite{alahi2016social}, and StarNet~\cite{zhu2019starnet}. Our dataset can boost the research of action and trajectory prediction in crowd scenes by involving more multimodal inputs or features.

\begin{table}[ht]
\caption{Results on trajectory prediction task. }
\centering
\begin{tabular}{lllll}
\toprule
        Methods          & FDE $\downarrow$ & MDE $\downarrow$ \\\toprule
LSTM~\cite{greff2016lstm}      & 1.133     &  0.648 \\\hline
Social-LSTM~\cite{alahi2016social} & 1.122 & 0.638       \\\hline
StarNet~\cite{zhu2019starnet} &0.983&0.404 \\\bottomrule
\end{tabular}
\label{predicition result}
\vspace{-2ex}
\end{table}

\subsection{More Applications}
Accurate pedestrian perception leads to wider applications, like the on-campus delivery robots and intelligent patrols for stations, for which dense and crowded campus scene would be the major challenges. Our dataset can provide a benchmark and challenging metrics for them. 

\section{Conclusion}
Focusing on 3D perceptions in crowded scenarios, we propose a new multimodal dataset with diverse crowd densities, multiple scenes, various weather, and different human poses. In particular, our dataset contain situations with high density and severe occlusions, which is challenging for current 3D perception methods. Based on multimodal data and annotations, our dataset can facilitate many perception tasks. Benchmarks on most of the tasks are provided in the paper. In addition, we propose a novel method to achieve more accurate perception on crowded scenes by considering the properties of pedestrian distribution. Experiments illustrate the superiority and generalization capability of our method.

{\small
\bibliographystyle{ieee_fullname}
\bibliography{egbib}
}

\end{document}